\documentclass{article} 
\usepackage{iclr2027_conference,times}

\usepackage{amsmath,amsfonts,bm}

\def\eqref#1{equation~\ref{#1}}

\def\1{\bm{1}}

\DeclareMathAlphabet{\mathsfit}{\encodingdefault}{\sfdefault}{m}{sl}
\SetMathAlphabet{\mathsfit}{bold}{\encodingdefault}{\sfdefault}{bx}{n}

\usepackage{hyperref}
\usepackage{url}
\usepackage{graphicx} 
\usepackage{algorithm}
\usepackage{algorithmic}
\usepackage{natbib}
\usepackage{amsmath}
\usepackage{amssymb}
\usepackage{amsthm}
\usepackage[utf8]{inputenc} 
\usepackage[T1]{fontenc}    
\usepackage{hyperref}       
\usepackage{url}            
\usepackage{booktabs}       
\usepackage{amsfonts}       
\usepackage{nicefrac}       
\usepackage{booktabs}
\usepackage{threeparttable}
\usepackage[table]{xcolor}
\usepackage{adjustbox}
\usepackage{microtype}      
\usepackage{xcolor}         
\usepackage{xr}
\usepackage{wrapfig}
\usepackage{graphicx}
\usepackage{caption}
\usepackage{multirow}
\usepackage{appendix}

\definecolor{headerblue}{HTML}{EAF2F8}
\definecolor{bestgreen}{HTML}{EAF7EA}
\newcommand{\best}[1]{\cellcolor{bestgreen}\textbf{#1}}
\title{Latent Reasoning Does Not Lack Signal, \\It Lacks Invariance: Subspace-Constrained \\Refinement for Robust Reasoning}

\author{%
\begin{minipage}[t]{\dimexpr\textwidth-2\tabcolsep\relax}
\centering
\normalfont
\textbf{Arun Vignesh Malarkkan}\textsuperscript{1}%
\thanks{Corresponding author: \texttt{arun.malarkkan@asu.edu}.},
\textbf{Manan Roy Choudhury}\textsuperscript{1},
\textbf{Utkarsh Byahut}\textsuperscript{1}\\[3pt]
\textbf{Yash Ravindra Charde}\textsuperscript{1},
\textbf{Vivek Gupta}\textsuperscript{1},
\textbf{Xiquan Cui}\textsuperscript{2}\\[3pt]
\textbf{Xiangliang Zhang}\textsuperscript{3},
\textbf{Yanjie Fu}\textsuperscript{1}\\[8pt]
\textsuperscript{1}Arizona State University, Tempe, AZ, USA\\
\textsuperscript{2}Workday, Atlanta, GA, USA\\
\textsuperscript{3}University of Notre Dame, Notre Dame, IN, USA
\end{minipage}%
}

\iclrfinalcopy 
\begin{document}

\maketitle
\fancyhead{}

\begin{abstract}
\label{sec:abstract}

Latent reasoning models such as Coconut~\cite{hao2024training} perform multi-step inference in embedding space, avoiding the token overhead of Chain-of-Thought while preserving reasoning capability. Recent training-free refinement methods on latent reasoning models improve accuracy by steering latent trajectories using contrastive signals from stronger and weaker reference checkpoints. We show that these gains are structurally fragile along three axes:
(i) \textbf{hyperparameter sensitivity}, where fixed step sizes and memory rates exhibit task-dependent optima;
(ii) \textbf{reference-pair dependence}, where contrastive directions reflect pair-specific artifacts rather than stable reasoning signals; and 
(iii) \textbf{input sensitivity}, where updates in the full embedding space allow surface-level variations to induce divergent trajectories.
We argue that reliable latent refinement requires operating on trajectory-level invariances rather than unconstrained embedding directions. We propose \textbf{Trajectory-Invariant Latent Refinement (TILR)}, a training-free framework with two components. First, we identify a spectrally-concentrated refinement subspace in which reference models consistently differ across a calibration set, and project contrastive updates onto this subspace to filter unstable variation. 
Second, we apply an adaptive gating mechanism that scales update magnitude according to the alignment between the current signal and the learned subspace, suppressing unreliable corrections and recovering the unmodified backbone when alignment is low. 
Across six reasoning benchmarks, TILR consistently improves over prior refinement methods while substantially reducing sensitivity to input perturbations and reference selections. In particular, TILR improves answer agreement under semantically equivalent paraphrasing by $11.4\%$ and reduces trajectory variance by $\sim30\%$ relative to the existing refinement baselines on Coconut backbone, while maintaining comparable accuracy. These results demonstrate that constraining refinement to invariant subspaces yields more reliable latent reasoning without requiring retraining or backpropagation.
\end{abstract}

\vspace{-0.3cm}
\section{Introduction}
\label{sec:Intro}
Chain-of-Thought (CoT) prompting \citep{wei2022chain} improves reasoning in large language models by explicitly generating intermediate steps, but incurs substantial token overhead and produces trajectories that cannot be revised once generated. Latent reasoning methods instead perform multi-step inference directly in embedding space, iteratively updating a hidden state without emitting intermediate tokens. This approach reduces generation cost while retaining competitive reasoning performance across mathematical and multi-hop tasks. This shift to embedding-level reasoning raises a question: \textit{which directions in embedding space should a post-hoc correction act on, and when should it act at all?}

We observed that once a latent trajectory deviates from a correct reasoning path, the model has no explicit means to recover~\citep{hao2024training}. Recent work~\citep{wang2026efficient} addresses this by introducing training-free refinement methods that steer latent trajectories using contrastive signals derived from stronger and weaker reference checkpoints. These methods improve accuracy without modifying model parameters, making them attractive for deployment. However, we show that these gains are structurally fragile. We identify three recurring failure modes. First, \textit{hyperparameter sensitivity}: the refinement depends on fixed step sizes and memory rates whose optimal values vary across tasks and inputs. Second, \textit{reference-pair dependence}: the contrastive direction derived from a single checkpoint pair can reflect pair-specific artifacts rather than general reasoning quality. Third, \textit{input sensitivity}: because updates are applied in the full embedding space, semantically equivalent inputs (e.g., paraphrases or reordered options) can produce divergent latent trajectories.

\begin{wrapfigure}{r}{0.4\textwidth}
    \vspace{-20pt} 
    \centering
    \includegraphics[width=0.38\textwidth]{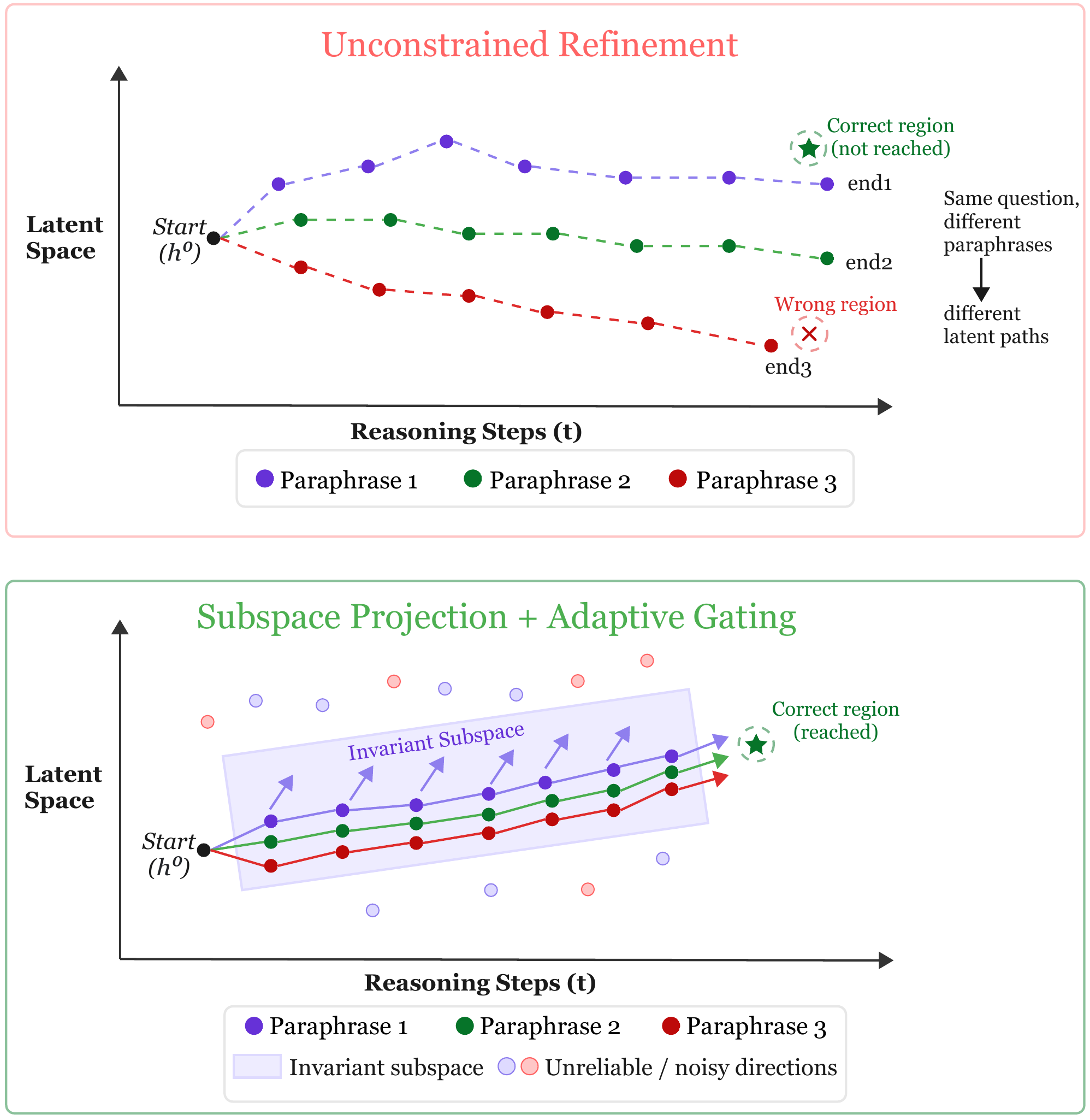}
    \caption{\textbf{Refinement vs TILR}}
    \vspace{-10pt}
    \vspace{-10pt}
    \label{fig:motivation}
\end{wrapfigure}

\textbf{Our Perspective.} We argue that these failure modes can be largely attributed to a common structural issue. The existing refinement methods treat the contrastive signal as a monolithic vector in the full embedding space, without distinguishing between components that are stable across inputs and reference choices and those that are instance-specific or spurious. As a result, unstable components are propagated into every refinement step, which leads to sensitivity along all three axes. 
The fix needs two components, organized around one geometric principle: (i) identify the subspace where refinement should operate only along directions that consistently encode reasoning-quality differences; (ii) apply corrections when the contrastive signal aligns with these stable directions, and suppressed otherwise.

\textbf{Proposed Framework.} We introduce \textbf{Trajectory-Invariant Latent Refinement (TILR)}, a training-free framework with two components. First, \textit{invariant subspace identification} extracts a spectrally concentrated refinement subspace in which strong and weak reference models consistently differ across a calibration set. Projecting contrastive updates onto this subspace filters out components that vary across inputs and reference pairs. Second, an \textit{adaptive stability gate} scales the update magnitude based on the alignment between the current contrastive signal and the learned subspace. When the alignment is high, the gate permits the full projected correction and when alignment is low, the update is suppressed and the method recovers the unmodified backbone, providing a principled fallback. 
Both components require only forward computation and add negligible inference overhead.

To evaluate refinement methods beyond average accuracy, we introduce a robustness-focused protocol that measures worst-case performance over hyperparameter neighborhoods, answer agreement under semantically equivalent input perturbations, and sensitivity to reference checkpoint choice.
Across six reasoning benchmarks, TILR preserves the accuracy gains of prior refinement methods while improving robustness to input perturbations and reference selection on most datasets.
\paragraph{Contributions.}
\begin{itemize}
    \item We identify three systematic fragility sources in latent reasoning refinement: hyperparameter sensitivity, reference dependence, and input sensitivity, and argue that they arise from a common failure to distinguish stable and unstable components of the contrastive signal.
    \item We propose TILR, a training-free refinement framework that constrains updates to an invariant subspace and modulates their magnitude based on signal reliability via invariant subspace projection and adaptive stability gating.
    \item We introduce a robustness evaluation protocol for latent reasoning that complements standard accuracy and empirically demonstrate that TILR improves robustness across multiple benchmarks while maintaining competitive accuracy and negligible computational overhead.
\end{itemize}

\clearpage
\vspace{-0.3cm}
\section{Related Work}
\label{sec:related}
\vspace{-0.2cm}
\paragraph{Latent reasoning.} Chain-of-Thought prompting~\citep{wei2022chain} proved the value of intermediate reasoning steps for LLMs, but each step consumes output tokens. Coconut \citep{hao2024training} reformulates reasoning as a recursive update in embedding space, eliminating the per-step token cost while matching CoT accuracy. This includes methods that learn to generate discrete reasoning paths in latent space \citep{deng2023implicit} and methods to internalize reasoning within hidden states during training \citep{deng2024explicit}. TILR operates on the latent reasoning paradigm and is agnostic to how the backbone was trained; it requires only that the backbone produce a sequence of latent states that can be intervened on at inference time.
\vspace{-0.2cm}
\paragraph{Post-training inference-time refinement.} Recent work modifies LLM behavior at inference time without parameter updates. Contrastive decoding \citep{li2023contrastive} and its derivatives steer token-level generation by combining logits from stronger and weaker model variants. Activation steering \citep{turner2023activation, rimsky2024steering} edits intermediate representations along directions identified from contrastive examples, for alignment rather than reasoning quality. The recent work on post-refinement \citep{wang2026efficient} adapts the contrastive principle to latent reasoning, using the difference between stronger and weaker checkpoints to steer Coconut's continuous thoughts. AdaAnchor \citep{sheshanarayana2026thinking} introduces adaptive halting for latent reasoning, terminating refinement when anchor representations stabilize. Where prior methods accept the contrastive signal as given, TILR decomposes it into stable and unstable components and applies only the stable part.
\vspace{-0.2cm}
\paragraph{Subspace projection and invariance.} Identifying low-rank structure in neural representations in domain adaptation \citep{ben2010theory} and continual learning \citep{saha2021gradient, lin2022trgp} are well-established, where projecting gradient updates onto subspaces preserves prior knowledge or filters distribution-specific variation. Invariant Risk Minimization \citep{arjovsky2019invariant} and the invariant prediction framework \citep{peters2016causal} formalize the principle that features whose predictive relationship is stable across environments generalize better than features merely correlated with the target. TILR draws on this principle without inheriting the optimization overhead: because the contrastive difference is directly observed via forward passes through reference models, the invariant directions can be extracted in closed form from a single SVD rather than learned through adversarial or bilevel optimization.

\section{Background and Problem Setup}
\label{sec:problem}


\paragraph{Latent Reasoning - Coconut backbone.} Latent reasoning models perform multi-step inference by iteratively updating a hidden state in embedding space rather than generating explicit intermediate tokens~\citep{hao2024training}. Let $f$ denote a frozen decoder block and $h_t \in \mathbb{R}^d$ the latent reasoning state at step $t$. The trajectory evolves as
\[
h_t = f(h_{t-1}), \quad t = 1, \dots, T.
\]

which provides no mechanism for error correction once the trajectory diverges from a correct reasoning path. 

\paragraph{Post-training latent refinement.} Recent work introduces training-free refinement mechanisms that modify the latent trajectory at inference time. A common approach combines residual blending with contrastive steering using a pair of reference models $(f_{\text{good}}, f_{\text{bad}})$:
\[
\tilde{h}_t = \alpha h_{t-1} + (1 - \alpha) f(h_{t-1}),
\]
\[
h_t = \tilde{h}_t + \eta \, d_t,
\quad \text{where} \quad d_t = h_t^{\text{good}} - h_t^{\text{bad}}.
\]
Here $h_t^{\text{good}} = f_{\text{good}}(\tilde{h}_t)$ and $h_t^{\text{bad}} = f_{\text{bad}}(\tilde{h}_t)$. Under the stopped-gradient approximation, $d_t$ depends only on the reference pair and the current input.
This refinement improves accuracy without parameter updates, but introduces a key structural limitation: the same contrastive direction is applied uniformly across all inputs and steps.

\paragraph{Observed fragility.}
Empirically, this leads to three recurring failure modes:
(i) sensitivity to hyperparameters $(\alpha, \eta)$,
(ii) dependence on the specific reference checkpoint pair, and
(iii) divergence under semantically equivalent input reformulations.

\paragraph{Problem formulation.}
Given a frozen coconut backbone $f$ and a reference pair $(f_{\text{good}}, f_{\text{bad}})$, we seek a refinement operator
\[
h_t = \mathcal{R}(\tilde{h}_t, h_t^{\text{good}}, h_t^{\text{bad}})
\]
that improves reasoning performance while being robust to:
(i) hyperparameter variation,
(ii) reference checkpoint choice, and
(iii) input reformulation.
The key constraint is that $\mathcal{R}$ must operate without parameter updates or backpropagation through the reference models.

\section{Trajectory-Invariant Latent Refinement (TILR)}
\label{sec:method}

TILR is motivated by a single observation: the contrastive direction $d_t$ that drives latent-reasoning refinement is not a homogeneous signal, but a mixture of components that are stable across inputs and reference choices, and components that are instance-specific and noisy. Existing methods apply $d_t$ uniformly in the full embedding space, allowing unstable components to propagate into every update.
TILR addresses this gap by separating these components and applying only the stable part of the signal, with magnitude determined by its reliability. Concretely, TILR adds two components to the refinement contrastive signal: an invariant subspace that constrains where the correction can act, and an adaptive gate that controls how strongly it acts. The subspace operationalizes the decomposition; the gate provides the fallback.

\begin{figure}[t]
\centering
\includegraphics[width=0.9\textwidth, height=4.5cm]{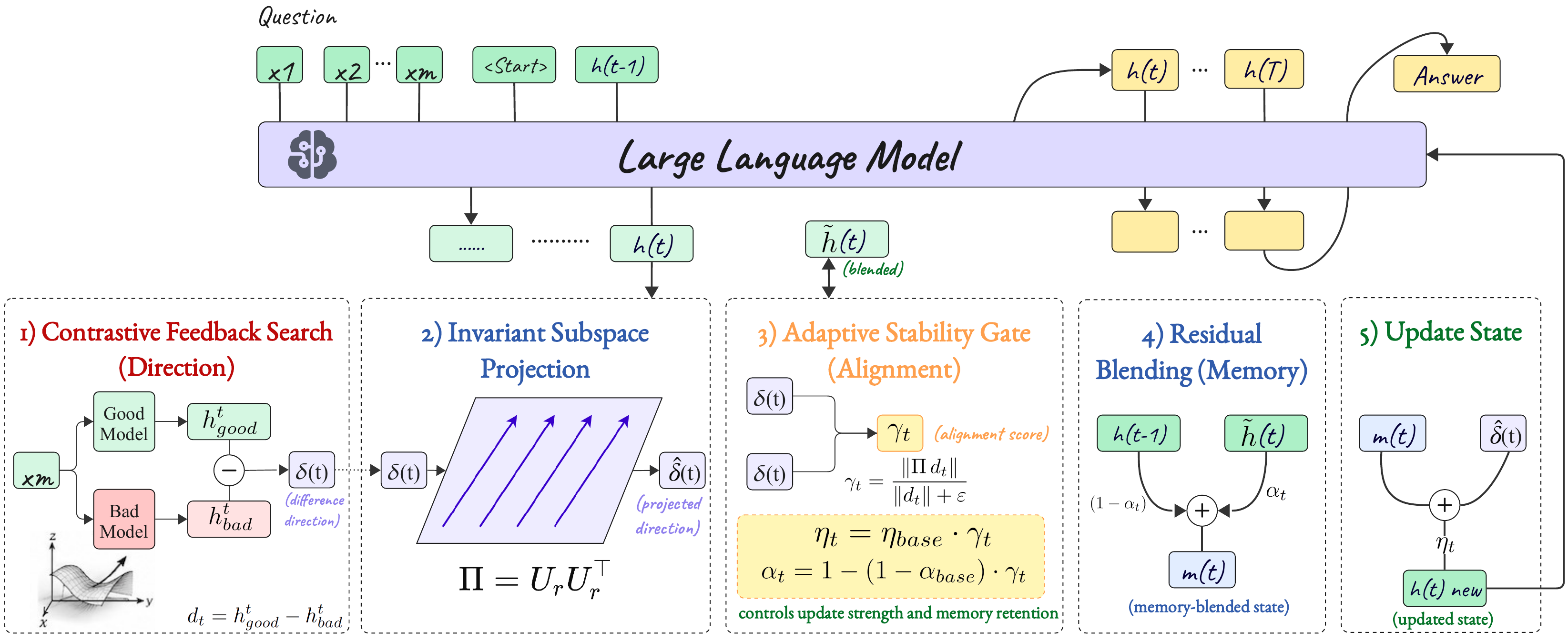}
\caption{\textbf{The Trajectory-Invariant Latent Refinement (TILR) Framework Overview.} TILR enhances latent reasoning by constraining contrastive updates to a stable reasoning subspace. (1) \textbf{Contrastive Feedback Search} identifies a direction $d_t$ pointing from weak to strong reference models. (2) \textbf{Invariant Subspace Projection} filters out instance-specific noise by projecting $d_t$ onto a calibrated subspace $U_r$. (3) \textbf{Adaptive Stability Gating} computes an alignment score $\gamma^t$ to modulate the update strength. (4) \textbf{Residual Blending} maintains trajectory stability, and (5) \textbf{State Updating} produces the refined latent state $h^{t}_{new}$ for the next reasoning step.}
\label{fig:tilr-methodology}
\vspace{-0.5cm}
\end{figure}

\paragraph{Empirical Hypothesis: Spectral Concentration}

Let $\delta_t = h_t^{\text{good}} - h_t^{\text{bad}}$ denote the contrastive difference. Empirically, we observe substantial concentration of contrastive energy in the leading spectral directions across inputs and reasoning steps.

We therefore posit that there exists a subspace $S \subset \mathbb{R}^d$ such that
\[
\mathbb{E}[\|\Pi_S^\perp \delta_t\|^2] \ll \mathbb{E}[\|\delta_t\|^2],
\]
where $\Pi_S^\perp$ is the orthogonal complement projection. This is validated empirically in Section~\ref{sec:experiments}.


\subsection{Invariant Subspace-Projected Contrastive Refinement}
\subsubsection{Invariant Subspace Identification}
\label{sec:subspace}

Given a calibration set $\mathcal{C} = \{x_1, \ldots, x_N\}$ of modest size $N$, we construct the subspace via forward passes only. 
We draw $\mathcal{C}$ uniformly at random from the training split of each benchmark with $N{=}200$ and a fixed seed, disjoint from the test set. The procedure uses only the contrastive differences $\delta^t_i$ from forward passes through the reference models and requires no labels; full construction details are in Appendix~\ref{app:calibration}.
For each $x_i$, we run the Coconut backbone with residual blending (Section~\ref{sec:problem}) to obtain $\{\tilde{h}^t_i\}_{t=1}^T$, we pass each $\tilde{h}^t_i$ through both reference models, and form the contrastive differences
\begin{equation}
    \delta^t_i = h^t_{\text{good},i} - h^t_{\text{bad},i} \in \mathbb{R}^d,
    \quad i = 1, \ldots, N, \quad t = 1, \ldots, T.
    \label{eq:delta}
\end{equation}
Stacking all $NT$ differences column-wise gives
\begin{equation}
    \Delta = \left[\delta^1_1,\; \ldots,\; \delta^T_1,\; \delta^1_2,\; \ldots,\; \delta^T_N\right]
    \in \mathbb{R}^{d \times NT}.
    \label{eq:delta-matrix}
\end{equation}
We compute the truncated SVD
\begin{equation}
    \Delta \approx U_r \Sigma_r V_r^\top,
    \quad U_r \in \mathbb{R}^{d \times r},
    \quad r = \min\Big\{k : \textstyle\sum_{j=1}^k \sigma_j^2 \,\big/\, \sum_{j=1}^d \sigma_j^2 \geq \tau\Big\},
    \label{eq:svd}
\end{equation}
where $r$ is chosen to retain a fraction $\tau$ of the variance. The invariant subspace $\mathcal{S}_r$ is defined by the projection 
\begin{equation}
    \Pi = U_r U_r^\top \in \mathbb{R}^{d \times d}.
    \label{eq:projection}
\end{equation}

The columns of $U_r$ are the principal components of the \emph{difference} between strong and weak reasoning outputs, not of the embedding space itself. This distinction matters: PCA on raw embeddings would capture directions of maximal input variance (dominated by lexical and stylistic variation), while our procedure captures directions of maximal reasoning-quality variance. We measure the geometric overlap between these two bases by computing the mean principal angle; a large angle confirms that the subspace captures structure distinct from generic input variation. Pooling $\delta^t_i$ across steps implements the spectral structure hypothesis; we ablate against per-step subspaces in the appendix.

\vspace{-6pt}
\subsubsection{Subspace-Projected Contrastive Refinement}
\label{sec:projected-refinement}

At inference, we replace the unconstrained update with its subspace-projected counterpart. The per-instance contrastive direction is
\begin{equation}
    d_t \triangleq h^t_{\text{good}} - h^t_{\text{bad}} \in \mathbb{R}^d,
    \label{eq:direction}
\end{equation}
and the projected refinement is
\begin{equation}
    h^t = \tilde{h}^t + \eta_t \cdot \Pi\, d_t,
    \label{eq:projected-update}
\end{equation}

where $\eta_t$ is the adaptive step size defined next. The projection $\Pi d_t$ retains only the component of $d_t$ aligned with the calibrated reasoning-quality subspace, attenuating instance-specific and pair-specific variation to the extent the hypothesis holds. This addresses reference invariance and input invariance: the components of $d_t$ responsible for these two fragility sources lie outside $\mathcal{S}_r$ by the assumption and are filtered out by $\Pi$.

\vspace{-3pt}
\subsection{Adaptive Stability Gate}
\label{sec:gate}

The projection controls where the update acts; the gate controls how strongly. We define the \textbf{subspace alignment score}
\begin{equation}
    \gamma_t = \frac{\|\Pi\, d_t\|}{\|d_t\| + \varepsilon},
    \label{eq:gamma}
\end{equation}
with $\varepsilon$ controlling  numerical stability. $\gamma_t \in [0, 1]$ measures the fraction of the current correction signal that lies within $\mathcal{S}_r$: $\gamma_t \approx 1$ indicates the signal is well-aligned with historically consistent quality-difference directions; $\gamma_t \approx 0$ indicates the signal is dominated by instance-specific components. We use $\gamma_t$ to scale the step size:
\begin{equation}
    \eta_t = min(\eta_{\text{base}} \cdot \gamma_t^2, 1).
    \label{eq:adaptive-eta}
\end{equation}
The base mixing rate $\alpha_{\text{base}}$ is \emph{not} gated.
The gate interpolates smoothly between two regimes:
\begin{equation}
    h^t \;\longrightarrow\;
    \begin{cases}
        \tilde{h}^t + \eta_{\text{base}} \cdot \Pi d_t & \text{as } \gamma_t \to 1 \quad \text{(full projected correction)},\\[4pt]
        \alpha_{\text{base}} h^{t-1} + (1-\alpha_{\text{base}}) f(h^{t-1}) & \text{as } \gamma_t \to 0 \quad \text{(uncorrected backbone)}.
    \end{cases}
    \label{eq:gate-regimes}
\end{equation}
When $\gamma_t$ is high, the full projected correction is applied on top of
the residual blend. As $\gamma_t \rightarrow 0$, the correction vanishes and
TILR recovers the uncorrected system.

\noindent \textbf{Sensitivity flattening on low-confidence inputs}. A direct consequence of the multiplicative gate
$\eta_t = min(\eta_{\text{base}} \cdot \gamma_t^2, 1)$ is that inputs with small $\gamma_t$ become approximately insensitive to the choice of $\eta_{\text{base}}$: scaling $\eta_{\text{base}}$ by any factor scales a near-zero quantity by that factor, leaving the effective correction negligible. This provides a plausible mechanism by which the gate addresses fragility source~(i): low-$\gamma$ inputs contribute disproportionately to the worst-case sensitivity region in the $(\alpha, \eta)$ grid, and TILR suppresses them. A local sensitivity analysis of the adaptive gating mechanism is provided in Appendix~\ref{app:sensitivity_theory}, showing that low-alignment inputs induce suppressed update sensitivity with respect to perturbations in the nominal step size. 

\noindent \textbf{Why $\alpha$ is not gated.} TILR is a post-training add-on, and the appropriate ``do no harm'' baseline is the uncorrected system at $\alpha = \alpha_{\text{base}}$. With a fixed $\alpha$, the limit $\gamma_t \to 0$ recovers this baseline exactly: the correction vanishes and ~\eqref{eq:projected-update} collapses to the residual blend. Modulating $\alpha_t$ with $\gamma_t$ would instead push the limit to $h^t = h^{t-1}$ (state freeze), which is not the original system's behavior and may be strictly worse. More fundamentally, $\alpha_{\text{base}}$ controls the backbone's memory-vs-computation balance, a property of the model dynamics; $\gamma_t$ measures the reliability of the correction direction and carries no information about whether $f(h^{t-1})$ itself is trustworthy. Modulating $\alpha$ by $\gamma_t$ conflates these two independent uncertainties. 

Overall, TILR decomposes the contrastive signal as
\[
d_t = \underbrace{\Pi d_t}_{\text{stable}} + \underbrace{(I - \Pi)d_t}_{\text{unstable}},
\]
and applies only the stable component, scaled by its reliability. This directly addresses input and reference sensitivity, while the gating mechanism limits the impact of unreliable corrections.

\noindent \textbf{Computational Overhead.}
At each reasoning step, TILR requires two matrix-vector products to compute $U_r^\top d_t$ and $U_r(U_r^\top d_t)$, with total complexity $\mathcal{O}(dr)$.
For the GPT-2 backbone ($d=768$), the learned subspace ranks range from $r=38$ to $171$ across the evaluated benchmarks. In practice, these operations add less than $3\%$ wall-clock inference overhead relative to unconstrained refinement. Calibration is performed once per benchmark and reference pair and is dominated by the reference forward passes and a truncated SVD of the $d\times NT$ contrastive-difference matrix.

\vspace{-0.2cm}
 \section{Experimental Setup}
\label{sec:experiments}
We design experiments to evaluate whether TILR improves the reliability of latent reasoning refinement while preserving accuracy. In particular, we test five questions:
\textbf{RQ1 (Accuracy):} Does TILR improve reasoning accuracy relative to existing latent refinement methods?
\textbf{RQ2 (Reference Robustness):} Is TILR robust to the choice of contrastive reference checkpoint pair?
\textbf{RQ3 (Input Invariance):} Does TILR produce more consistent predictions
and latent trajectories to semantically equivalent input reformulations?
\textbf{RQ4 (Subspace Geometry):} Do contrastive refinement signals exhibit a spectrally concentrated subspace distinct from generic representation variance?
\textbf{RQ5 (Causal Role of Subspace):} Is the learned subspace functionally necessary for refinement performance?
Additional ablations, hyperparameters and cross architecture generalization analysis are reported in the Appendix.
\noindent \textbf{Benchmarks.} We evaluate on six reasoning benchmarks covering arithmetic, robustness-oriented mathematical reasoning, and symbolic deduction. \textbf{GSM8K-Aug}
\citep{deng2023implicit} serves as the standard arithmetic reasoning benchmark.
\textbf{SVAMP} \citep{patel2021nlp} evaluates robustness to lexical and structural
variation in arithmetic word problems, while \textbf{GSM-Plus}
\citep{li2024gsm} introduces controlled numerical and structural perturbations
of GSM8K problems. We additionally evaluate on \textbf{GSM-Symbolic}
\citep{mirzadeh2025gsm}, which generates symbolic variants of GSM8K-style
problems, and on the deductive reasoning benchmarks \textbf{ProsQA}
\citep{hao2024training} and \textbf{ProntoQA}
\citep{saparov2022language}. This combination of benchmarks allows us to evaluate
both standard reasoning performance and robustness under controlled changes to
the problem representations.

\noindent \textbf{Models and checkpoints.} The primary backbone is \textbf{GPT-2 base (117M)} trained with the Coconut latent reasoning framework~\citep{hao2024training}. A separate backbone is trained per benchmark to avoid confounding refinement quality with cross-task transfer. Reference models $(f_{\text{good}}, f_{\text{bad}})$ are selected from different training epochs of the same backbone, with $f_{\text{good}}$ achieving higher validation accuracy. Unless otherwise specified, we use later-stage and early-stage checkpoints to construct the contrastive pair. To test whether the findings hold when scaled, we additionally evaluate
\textbf{Llama-3.2-1B} \citep{grattafiori2024llama3herdmodels} and \textbf{Qwen2.5-Math-1.5B} \citep{yang2024qwen25mathtechnicalreportmathematical} in the Appendix. No checkpoint, hyperparameter, or subspace is selected using any test split. 



\noindent \textbf{Baselines.} We compare TILR against five methods under identical conditions: (i)~\textbf{No-CoT}~\citep{radford2019language}: Direct prediction without reasoning
steps; 
(ii)~\textbf{CoT}: Token-based chain-of-thought reasoning with a separately trained model; (iii)~\textbf{Coconut}~\citep{hao2024training}: Latent backbone without refinement;
(iv)~\textbf{Coconut + Refinement} \citep{wang2026efficient}: Unconstrained  contrastive latent refinement; and
(v)~\textbf{AdaAnchor} \citep{sheshanarayana2026thinking}: Adaptive anchor refinement with step-level adaptivity. 
Except CoT, all baselines operate on the same per-benchmark Coconut backbone, so observed differences trace to the refinement mechanism alone.

\noindent \textbf{Metrics.} 
Average accuracy alone does not capture the reliability of latent refinement. We therefore evaluate using four complementary metrics: \textbf{M1: Accuracy} is exact-match on the test set, averaged over 3 seeds.
\textbf{M2: Reference Sensitivity} is the variation in accuracy across different reference checkpoint pairs, measured via standard deviation and range.
\textbf{M3: Answer agreement} is the fraction of consistent predictions across multiple semantically equivalent reformulations of each input. 
\textbf{M4: Trajectory variance} is the mean pairwise $\ell_2$ distance between latent trajectories induced by input reformulations.
These metrics correspond directly to the fragility sources identified in Section 3.

\vspace{-0.3cm}
\section{Experimental Results}
\label{sec:results}

\noindent \textbf{RQ1: Overall Reasoning Performance.}
Across six reasoning benchmarks, we evaluate whether constraining latent refinement to the learned subspace improves reasoning accuracy without sacrificing the efficiency
advantage of latent reasoning. Table~\ref{tab:rq1-accuracy} shows that TILR outperforms latent refinement baselines on all six benchmarks, while preserving the token efficiency of latent reasoning.  CoT remains strongest on GSM8K-Aug and GSM-Plus, which is consistent with the known advantage of explicit symbolic traces on open-ended multi-step arithmetic reasoning. The comparison with explicit
CoT further illustrates the accuracy--efficiency trade-off. Overall, TILR improves the reliability of latent refinement within the same efficiency regime rather than closing the token-level CoT gap. AdaAnchor is below Coconut on most benchmarks in this setup, so we treat it as a secondary reference point rather than a strong competing baseline. We observe similar trends on a stronger Qwen2.5-Math-1.5B and Llama3.2-1B-Instruct backbones (Appendix~\ref{app:cross_arch}), where TILR improves over unconstrained refinement on all benchmarks. Notably, these results use a single hyperparameter configuration selected on
GSM8K-Aug and transferred unchanged to the remaining benchmarks, rather than
tuning TILR separately for each test domain.

\begin{table}[t]
\centering
\caption{Exact-match accuracy (\%) and average generation tokens per
inference on six reasoning benchmarks under GPT-2 backbones trained
per-benchmark.  Values are mean $\pm$ standard error.
\textbf{Bold} indicates best overall; $^{\dagger}$ best among latent methods;
\underline{underline} second-best method. Token counts are averaged across seeds.}
\vspace{0.1cm}
\label{tab:rq1-accuracy}
\small
\setlength{\tabcolsep}{6pt}
\renewcommand{\arraystretch}{1.0}

\begin{tabular}{lcccccc}
\toprule
\rowcolor{headerblue}
& \multicolumn{2}{c}{\textbf{GSM8K}} 
& \multicolumn{2}{c}{\textbf{SVAMP}} 
& \multicolumn{2}{c}{\textbf{GSM-Plus}} \\
\rowcolor{headerblue}
\textbf{Method} & \textbf{Acc.} & \textbf{Tok.} 
                & \textbf{Acc.} & \textbf{Tok.} 
                & \textbf{Acc.} & \textbf{Tok.} \\
\midrule
No-CoT     & 6.87\,$\pm$\,0.3 & $2.5$ & 23.20\,$\pm$\,1.33 & $2$ & 2.71\,$\pm$\,0.2 & $3$ \\
CoT               & \best{32.83\,$\pm$\,1.3} & $28$ & 28.20\,$\pm$\,1.5 & $24$ & \best{16.20\,$\pm$\,0.4} & $30$ \\
\midrule
Coconut        & 26.61\,$\pm$\,1.2 & $8$ & 30.00\,$\pm$\,0.8 & $4$ & 12.93\,$\pm$\,0.3 & $8$ \\
Refinement   & 26.72\,$\pm$\,1.1 & $8$ & \underline{31.50\,$\pm$\,1.4} & $4$ & 13.46\,$\pm$\,0.4 & $8$ \\
AdaAnchor  & 26.54\,$\pm$\,0.8 & $8$ & 27.10\,$\pm$\,1.5 & $4$ & 12.98\,$\pm$\,0.3 & $8$ \\
\midrule
\textbf{TILR}                 & \underline{27.67\,$\pm$\,1.2$^{\dagger}$} & $8$ & \best{33.70\,$\pm$\,1.4$^{\dagger}$} & $4$ & \underline{13.70\,$\pm$\,0.3$^{\dagger}$} & $8$ \\
\bottomrule
\end{tabular}

\vspace{0.1cm}

\begin{tabular}{lcccccc}
\toprule
\rowcolor{headerblue}
& \multicolumn{2}{c}{\textbf{GSM-Symbolic}} 
& \multicolumn{2}{c}{\textbf{ProsQA}} 
& \multicolumn{2}{c}{\textbf{ProntoQA}} \\
\rowcolor{headerblue}
\textbf{Method} & \textbf{Acc.} & \textbf{Tok.} 
                & \textbf{Acc.} & \textbf{Tok.} 
                & \textbf{Acc.} & \textbf{Tok.} \\
\midrule
No-CoT     & 4.32\,$\pm$\,1.1 & $3$ & 76.41\,$\pm$\,1.5 & $9$ & 82.22\,$\pm$\,0.88 & $3$ \\
CoT               & 10.83\,$\pm$\,0.8 & $22$ & 72.34\,$\pm$\,1.1 & $51$ & 92.20\,$\pm$\,0.6 & $85$ \\
\midrule
Coconut        & 10.86\,$\pm$\,0.6 & $9$ & 88.50\,$\pm$\,1.1 & $15$ & 92.93\,$\pm$\,1.2 & $9$ \\
Refinement   & \underline{10.87\,$\pm$\,0.5} & $8$ & \underline{88.53\,$\pm$\,0.4} & $16$ & 93.46\,$\pm$\,1.1 & $9$ \\
AdaAnchor  & 10.09\,$\pm$\,1.2 & $8$ & 87.10\,$\pm$\,0.1 & $16$ & \underline{92.98\,$\pm$\,0.3} & $9$ \\
\midrule
\textbf{TILR}                 & \best{11.72\,$\pm$\,0.6 $^{\dagger}$} & $8$ & \best{90.4\,$\pm$\,0.1$^{\dagger}$} & $16$ & \best{94.70\,$\pm$\,0.1$^{\dagger}$} & $9$ \\

\bottomrule
\end{tabular}
\vspace{-0.5cm}
\end{table}

\noindent \textbf{RQ2: Checkpoint-pair sensitivity.}
To isolate the effect of the contrastive pair, we evaluate six good/bad checkpoint pairs per dataset using Coconut backbones trained without stepwise internalization. For each pair, we report the standard deviation, and range across pairs. As shown in Table~\ref{tab:rq2_checkpoint_sensitivity}, TILR substantially reduces checkpoint-pair variability on GSM8K-Aug, ProsQA, ProntoQA, and GSM-Plus, with relative standard-deviation reductions of $72.7\%$, $60\%$, $64.2\%$, and $53.0\%$, respectively.  The reduction in both dispersion measures indicates that TILR's performance depends less strongly on the particular reference pair used at inference time. Importantly, the robustness improvement is observed across datasets with substantially different mean subspace overlap. Overall, these results provide empirical evidence that constraining contrastive refinement to the learned subspace makes
inference less sensitive to reference-checkpoint selection.

\begin{table}[t]
\centering
\small
\setlength{\tabcolsep}{6pt}
\renewcommand{\arraystretch}{0.7}
\caption{
\textbf{Robustness to checkpoint-pair selection.}
Lower values indicate stronger robustness.
}
\label{tab:rq2_checkpoint_sensitivity}
\vspace{-0.3cm}
\begin{tabular}{lrrrrr}
\toprule
\rowcolor{headerblue}
Dataset &
\multicolumn{3}{c}{Sensitivity across pairs (pp, lower is better)} &
\multicolumn{2}{c}{Subspace} \\
\rowcolor{headerblue}
 & Ref. std & TILR std & Std. red. &
 Range red. & Mean overlap \\
\cmidrule(lr){2-4}
\cmidrule(lr){5-6}
\midrule
GSM8K      & 0.160 & \best{0.044} & \best{+72.7} & \best{+66.7} & 0.837 \\
ProsQA     & 0.525 & \best{0.210} & \best{+60.0} & \best{+68.5} & 0.564 \\
ProntoQA   & 0.605 & \best{0.216} & \best{+64.2} & \best{+60.2} &  0.370 \\
GSM-Plus   & 0.350 & \best{0.165} & \best{+53.0} & \best{+43.5} & 0.988 \\
\bottomrule
\end{tabular}
\vspace{-0.3cm}
\end{table}

\begin{figure}[t]
\centering
\includegraphics[width=0.75\textwidth, height=3.5cm]{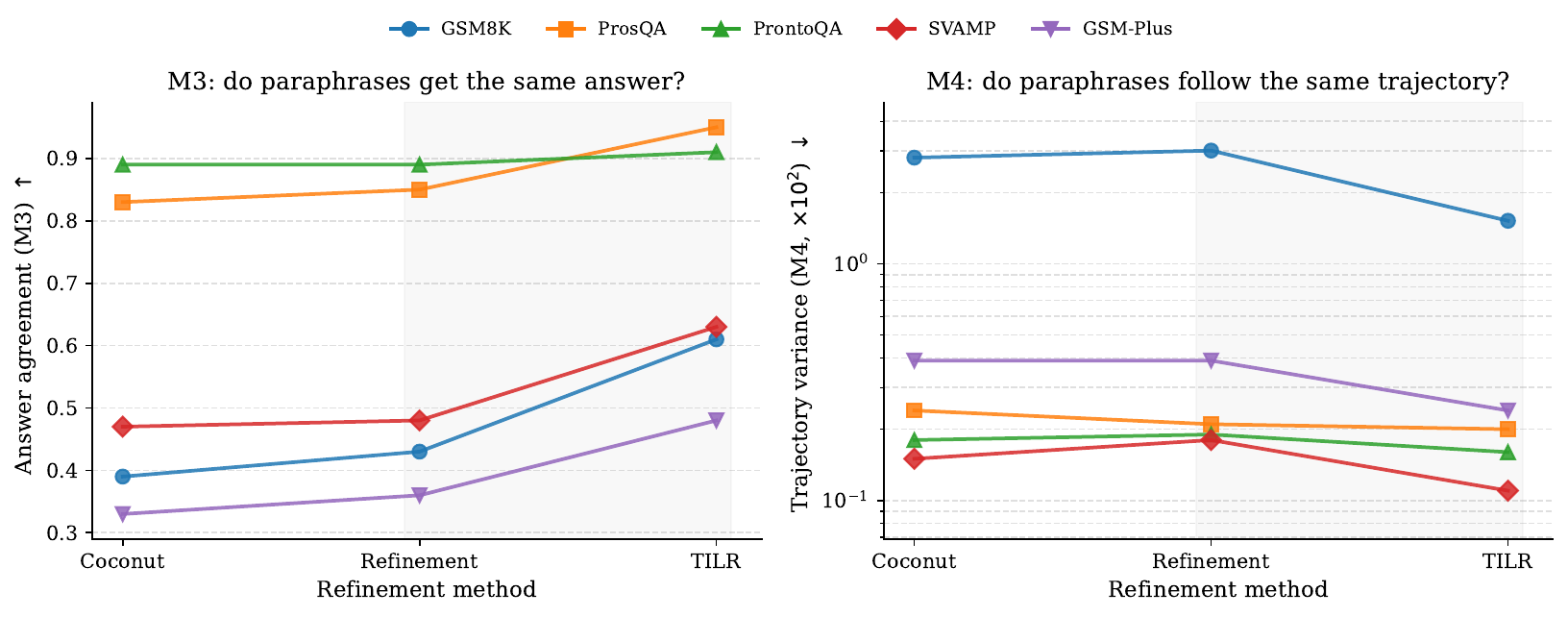}
\caption{\textbf{Input invariance under perturbations.}
\textbf{Left:} Answer agreement (M3, $\uparrow$) across semantically equivalent reformulations per input. \textbf{Right:} Latent trajectory
variance (M4, $\downarrow$).}
\label{fig:rq3-input-invariance}
\vspace{-0.5cm}
\end{figure}

\noindent \textbf{RQ3. Input Invariance}
\label{sec:rq3}
We next examine whether TILR produces more stable reasoning behavior under
semantically equivalent reformulations of the same input at two levels: Measuring answer agreement (M3, $\uparrow$), which captures
whether reformulations produce the same final prediction and latent trajectory variance (M4, $\downarrow$), which measures divergence between the
corresponding latent reasoning trajectories across $M{=}5$ semantically equivalent reformulations per test input generated using Llama-3-Instruct~\citep{grattafiori2024llama3herdmodels}  (Figure~\ref{fig:rq3-input-invariance}). We observe that TILR reduces trajectory variance relative to Refinement on every benchmark, with reductions of $49.5\%$ on GSM8K, $4.8\%$ on ProsQA, $15.8\%$ on ProntoQA, $38.9\%$ on
SVAMP, and $38.5\%$ on GSM-Plus, respectively. This supports the claim that subspace projection suppresses paraphrase-specific components of the contrastive direction before they affect the trajectory (Section~\ref{sec:complete}). The same pattern is observed in answer agreement. TILR improves answer agreement over unconstrained Refinement on all five evaluated benchmarks by an average of $\sim11.4\%$. These results show that TILR enables semantically equivalent inputs to be more likely to produce the same answer while following similar latent trajectories.

\noindent \textbf{RQ4: Spectral Geometry of the learned refinement subspace.}
Here, we study whether the contrastive refinement direction is concentrated in a low-dimensional spectral structure, and whether that structure is distinct from ordinary input-hidden-state variance. We measure the effective rank $r_{\tau}$ needed to explain $\tau$ fraction of
the contrastive covariance spectrum, the variance captured by the leading
directions, and the alignment between the learned contrastive basis and PCA
bases computed from input hidden states or latent refinement trajectories.
The input-controlled columns remove the input-PCA component before measuring
the residual contrastive signal.
Table~\ref{tab:rq4_subspace_geometry}, the signal is strongly concentrated in its leading directions: the first component explains $32.0$--$52.0\%$ of the variance, while the top 10 components capture $76.0$--$83.6\%$. However, the rank required to explain $90\%$ of the variance
varies substantially across datasets, from $38$ on SVAMP to $171$ on GSM8K,
indicating a concentrated leading spectrum followed by a task-dependent
long tail. The learned contrastive subspace also exhibits varying alignment with
input-PCA directions, ranging from $0.340$ to $0.804$, suggesting that the refinement signal contains substantial structured concentration. These results support a reduced-dimensional representation of the dominant refinement signal rather than a uniformly low-rank structure.

\begin{table}[t]
\centering
\small
\setlength{\tabcolsep}{3pt}
\renewcommand{\arraystretch}{0.9}
\caption{
\textbf{Spectral structure of the learned refinement subspace.}
Low $r_{0.90}$ and high explained variance indicate strong spectrally concentrated structure.
}
\label{tab:rq4_subspace_geometry}
\begin{tabular}{lrrrr}
\toprule
\rowcolor{headerblue}
Dataset &
$r_{0.90}$ &
Top-1 EV &
Top-10 EV &
Align(input) \\
\midrule
GSM8K      & 171  & 0.320 & 0.760 & 0.779 \\
ProsQA     & 66  & 0.432 & 0.836 & 0.620  \\
ProntoQA   & 84  & 0.520 & 0.810 & 0.691  \\
SVAMP      & 38  & 0.425 & 0.765 & 0.340  \\
GSM-Plus   & 156  & 0.402 & 0.781 & 0.804  \\
\bottomrule
\end{tabular}
\end{table}

\begin{figure*}[t]
\centering
\includegraphics[width=0.75\textwidth, height = 5cm]{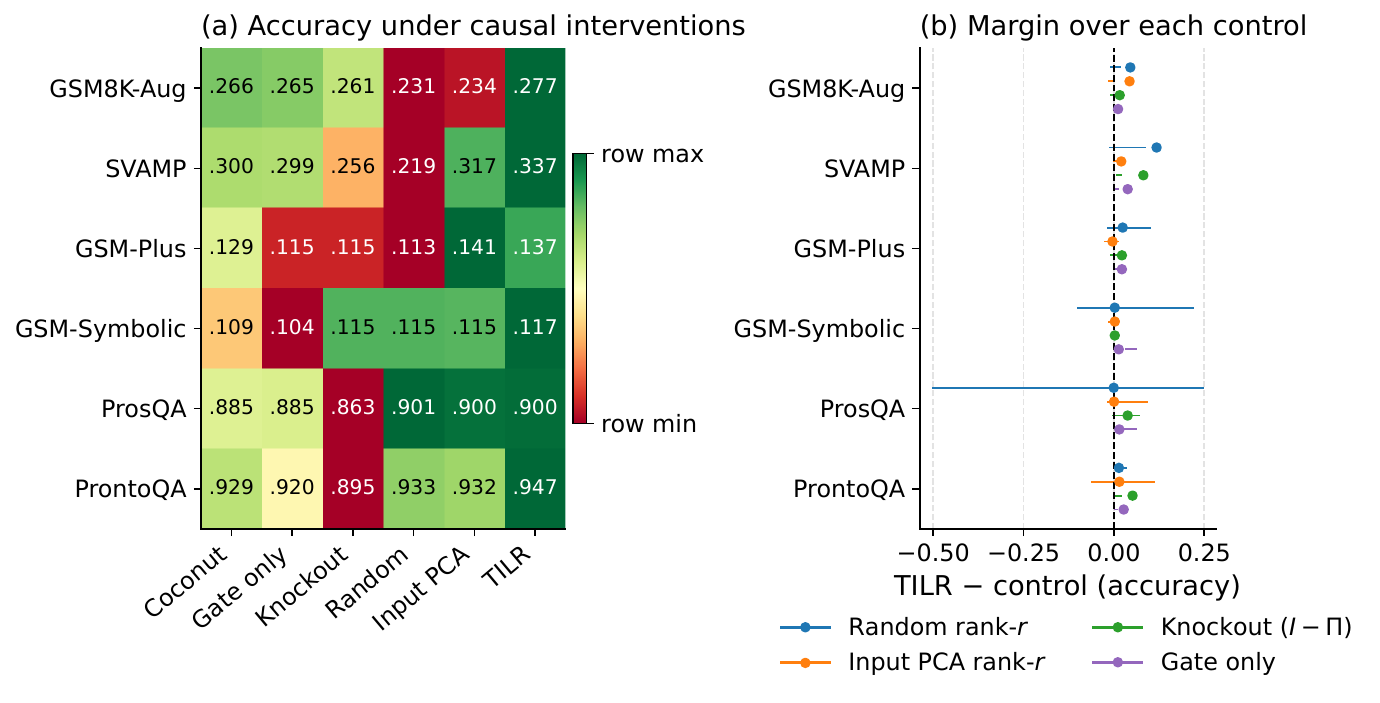}
\vspace{-0.2cm}
\caption{
\textbf{Causal subspace interventions.}
Panel (a) reports accuracy under different subspaces. Panel (b) shows TILR
margins over the strongest causal controls.
}
\label{fig:gpt2small_rq7_causal_interventions}
\end{figure*}

\noindent \textbf{RQ5: Functional Role of the Learned Subspace.}
We test whether the learned TILR subspace is causally useful rather than merely correlated with generic hidden-state variance. We use the same no-internalization Coconut backbones as in RQ2, we compare TILR against the base Coconut refinement trajectory, random subspaces, input/trajectory PCA subspaces, an orthogonal knockout intervention, and a raw-update gate baseline. The controlled intervention results~\ref{fig:gpt2small_rq7_causal_interventions} show that the learned subspace is consistently more useful than random or knocked-out directions, and the orthogonal knockout baseline is worse than TILR on five out of six datasets. This supports the interpretation that the learned directions carry task-relevant refinement information, not just low-rank structure.

Overall, TILR improves reasoning accuracy while making latent refinement more robust to reference selection and input reformulation. The learned subspace captures structured, functionally useful refinement directions, with additional sensitivity and ablation analyses provided in the Appendix.

\section{Conclusion}
\label{sec:conclusion}
We introduced TILR, a training-free refinement method for latent reasoning that separates the contrastive signal into subspace-aligned and residual components, applies updates only within a learned invariant refinement subspace, and modulates their magnitude based on alignment with that subspace. This formulation reframes latent refinement as a problem of selecting and trusting directions in embedding space, rather than uniformly applying contrastive signals.
Across six evaluated benchmarks, TILR improves accuracy over existing refinement methods while consistently reducing trajectory variance and improving answer consistency under input reformulations.
Our analysis shows that contrastive refinement signals are highly structured, with substantial variance concentrated in the leading directions between strong and weak reasoning trajectories. Controlled subspace intervention experiments further indicate that these directions are functionally useful for refinement, rather than incidental low-dimensional spectral structure. These results suggest that the key limitation of existing latent refinement methods is not the absence of useful signals, but the inability to distinguish stable structure from instance-specific variation. TILR addresses this by filtering and gating the contrastive signal, yielding more reliable behavior without retraining.
However, we also observed some limitations. The learned subspace is empirical rather than theoretically grounded, and its stability varies across datasets. In addition, while TILR improves input invariance, it does not fully eliminate trajectory divergence under reformulation. Understanding how invariant structure evolves across model scale and training regimes remains an open problem.
Our results point to a geometric perspective on latent reasoning, where the effectiveness of refinement depends on identifying and exploiting spectrally concentrated structure in representation space. Latent reasoning does not lack signal; it lacks invariance, and TILR fills the gap by selectively preserving the dominant refinement structure. Extending this perspective to other forms of inference-time control on larger models is a promising direction for future work.

\subsection*{AI use statement}

In this work, we used generative AI tools for debugging, standard code implementation, generation of evaluation data, and text polishing. Specifically, Llama-3-Instruct was used to generate the $M=5$ semantically equivalent input reformulations (paraphrases and lexical variations) used in the robustness evaluation protocol described in Sections 5 and 6. Commercial
LLM-based assistants were also used to support debugging and routine code implementation, as well as during the post-drafting phase to improve sentence structure and grammatical consistency. Generative AI tools were not used to develop the core research ideas, experimental design, or mathematical formulations. All AI-assisted code, evaluation data, and text were reviewed by the authors, including verification of generated reformulations for semantic equivalence to the original evaluation instances. The authors take responsibility for the final content, claims, experimental results, and artifacts presented in this work.

\subsection*{Ethics statement}

While Trajectory-Invariant Latent Refinement (TILR) is an algorithmic framework designed to improve the internal consistency of latent reasoning models, its downstream deployment carries considerations regarding reliability. TILR improves trajectory invariance under semantic perturbations rather than guaranteeing factual correctness. Consequently, systematic biases or errors encoded by the underlying reference checkpoints may persist under refinement, and improved consistency should not be interpreted as evidence of correctness or safety. Furthermore, because TILR improves inference-time robustness without requiring parameter retraining, it may facilitate the deployment of latent reasoning systems without addressing limitations inherited from the underlying models. We therefore view TILR as a robustness enhancement rather than a standalone verification or safety alignment solution. In safety-critical applications, it should be combined with task-specific verification and appropriate human oversight.

\subsection*{Reproducibility statement}

To ensure the reproducibility of our results, we have provided the code repository and comprehensive details of our framework across both the main text and supplementary materials. The complete step-by-step inference procedure for TILR is formalized in Algorithm 1 (Section 4). Detailed specifications of our calibration set construction, including sample sizes ($N=200$) and conditions for the truncated Singular Value Decomposition (SVD), are explicitly provided in Appendix E. Full theoretical derivations regarding the full Jacobian formulation are detailed in Appendix A. All baseline configurations, training epochs, and datasets (including GSM8K-Aug, SVAMP, GSM-Plus, GSM-Symbolic, ProsQA, and ProntoQA) are defined in Section 5. All experiments were executed on a hardware cluster comprising 4$\times$ NVIDIA H100 GPUs, requiring under 30 seconds for the one-time calibration phase as described in Appendix B. Code to replicate the calibration and inference steps will be made publicly available upon publication.
\clearpage



\bibliography{main}
\bibliographystyle{iclr2027_conference}

\clearpage
\appendix
\renewcommand{\appendixpagename}{\centering\Large Appendix}
\appendixpage

\section{Stopped-Gradient and the Full Jacobian}
\label{app:jacobian}

The simplification used in Section~\ref{sec:problem} treats $f_{\text{good}}(\tilde{h}^t)$ and
$f_{\text{bad}}(\tilde{h}^t)$ as constants with respect to $\tilde{h}^t$. The full gradient of the contrastive objective includes Jacobian terms from both reference models:
\begin{equation}
    \nabla_{\tilde{h}^t} \!\left[\operatorname{MSE}(h^t_{\text{good}}, \tilde{h}^t) - \operatorname{MSE}(h^t_{\text{bad}}, \tilde{h}^t)\right]
    = \tfrac{2}{d}(J_{\text{good}} - I)^\top (h^t_{\text{good}} - \tilde{h}^t)
    - \tfrac{2}{d}(J_{\text{bad}} - I)^\top (h^t_{\text{bad}} - \tilde{h}^t),
    \label{eq:full-gradient}
\end{equation}
where $J_{\text{good}}$ and $J_{\text{bad}}$ are the Jacobians at $\tilde{h}^t$.

We isolate this derivation in the appendix because the main paper only needs the stopped-gradient intuition. The full expression makes clear that the refinement update is not simply a static pull toward one reference and push away from another; it also depends on how both reference trajectories locally deform the hidden-state manifold. This is precisely why the projection step is useful. It suppresses directions that are unstable across reference choices before those Jacobian-induced effects can dominate the update.
\vspace{-0.3cm}

\section{Algorithm and Properties}
\label{sec:complete}

Combining the projection and gate, the full TILR update at each step is summarized in Algorithm~\ref{alg:tilr}, where $h^0$ denotes the initial latent state produced by the backbone's embedding of the input tokens. The computation proceeds sequentially with no circular dependency: $\alpha_{\text{base}}$ is fixed, $\gamma_t$ is computed once per step from the subspace projection, and the correction is applied additively.

Two design guarantees follow directly from the algorithm's structure, corresponding to the two failure modes the method was designed to handle. 
\paragraph{Graceful degradation.} As $\gamma_t \to 0$, the correction magnitude $\eta_t \cdot \Pi d_t$ vanishes and Step~4 reduces to
\begin{equation}
    h^t \;\longrightarrow\; \alpha_{\text{base}} \cdot h^{t-1} + (1-\alpha_{\text{base}}) \cdot f(h^{t-1}),
    \label{eq:graceful-degradation}
\end{equation}
which is exactly the uncorrected residual-blended backbone.

\paragraph{Subspace-mediated input invariance.} By construction, two inputs $x, x'$ whose contrastive directions differ only outside $\mathcal{S}_r$ satisfy $\Pi d_t(x) = \Pi d_t(x')$. If additionally
$\|d_t(x)\| = \|d_t(x')\|$, then $\gamma_t(x) = \gamma_t(x')$ and the TILR-induced divergence reduces to the backbone-induced divergence:
\begin{equation}
    h^t(x) - h^t(x') \;=\; \tilde{h}^t(x) - \tilde{h}^t(x').
    \label{eq:input-invariance}
\end{equation}
The correction contributes no additional divergence beyond that produced by the backbone forward pass. 
Though TILR is conceptually inspired by invariant prediction~\citep{peters2016causal, arjovsky2019invariant}, we make no formal claim that $\mathcal{S}_r$ corresponds to causally invariant features~\citep{peters2016causal}, only that empirically it captures directions along which reasoning-quality differences are consistent across the calibration population.

\begin{algorithm}[t]
\caption{TILR: Inference Procedure}
\label{alg:tilr}
\begin{algorithmic}[1]
\REQUIRE Input $x$, backbone $f$, references $f_{\text{good}}, f_{\text{bad}}$,
projection $\Pi$, hyperparameters $\eta_{\text{base}}, \alpha_{\text{base}}$,
steps $T$
\ENSURE Answer $y$
\STATE $h^0 \leftarrow $ initial latent state for input $x$
\FOR{$t = 1$ to $T$}
    \STATE $\tilde{h}^t \leftarrow \alpha_{\text{base}} h^{t-1} + (1 - \alpha_{\text{base}}) f(h^{t-1})$
    \hfill $\triangleright$ Residual blending
    \STATE $d_t \leftarrow f_{\text{good}}(\tilde{h}^t) - f_{\text{bad}}(\tilde{h}^t)$
    \hfill $\triangleright$ Contrastive direction
    \STATE $\gamma_t \leftarrow \|\Pi d_t\| / (\|d_t\| + \varepsilon)$;\quad
    $\eta_t \leftarrow min(\eta_{\text{base}} \cdot \gamma_t^2,1)$
    \hfill $\triangleright$ Gated step size
    \STATE $h^t \leftarrow \tilde{h}^t + \eta_t \cdot \Pi d_t$
    \hfill $\triangleright$ Projected correction
\ENDFOR
\STATE $y \leftarrow \text{Decode}(h^T)$
\RETURN $y$
\end{algorithmic}
\end{algorithm}

\paragraph{Compute and reproducibility.} All experiments are run on 4$\times$ NVIDIA H100 GPUs. Calibration takes $< 30$ seconds for GPT-2; TILR inference adds $< 3\%$ wall-clock overhead over unconstrained refinement. 

\vspace{-0.3cm}
\section{Formalisms of Robustness Desiderata}
\label{app:desiderata}

We formalize the three robustness properties referenced in Section~\ref{sec:problem} here.

\begin{description}
    \item[1: Hyperparameter stability.] For any $(\alpha', \eta')$ in a neighborhood $\mathcal{N}(\alpha, \eta)$
    of the nominal configuration, the accuracy degradation is bounded:
    $|\operatorname{Acc}(\alpha', \eta') - \operatorname{Acc}(\alpha, \eta)| \leq \epsilon$
    for a small $\epsilon > 0$.
    \item[2: Reference invariance.] The refinement direction is robust to the choice of reference pair: substituting a different pair of comparable CoT validation accuracy does not change the sign of the accuracy improvement over the unrefined backbone.
    \item[3: Input invariance.] For any two semantically equivalent inputs $x$ and $x'$ (paraphrases or option reorderings), the refined latent trajectories satisfy
    $\frac{1}{T}\sum_{t=1}^{T}\|h^t(x) - h^t(x')\| \leq \delta$ for a small $\delta > 0$.
\end{description}

The unconstrained refinement~\cite{wang2026efficient} satisfies none of these properties by construction: $\alpha$ and $\eta$ are fixed scalars chosen per task (violating D1); the contrastive direction $d_t$ is computed from a single reference pair with no mechanism to average over comparable pairs (violating D2); and $d_t$ operates in the full embedding $\mathbb{R}^d$ without any constraint that surface-level input variations be projected away (violating D3).

TILR (Section~\ref{sec:method}) addresses D2 and D3 directly via the subspace projection: components of $d_t$ that vary across reference pairs or across input reformulations are removed from the update by construction. D1 is addressed empirically rather than by construction. The gating mechanism scales the effective step size by the alignment score $\gamma_t$, which means that on low-confidence inputs (small $\gamma_t$) the effective $\eta$ is suppressed regardless of the nominal $\eta_{\text{base}}$. This provides a plausible reason for flatter sensitivity surfaces, but it is still an empirical claim rather than a theorem. For that reason, the appendix reports the complete RQ2 surfaces rather than only the tuned operating points.
\vspace{-0.3cm}
\section{Local Sensitivity Analysis of Adaptive Gating}
\label{app:sensitivity_theory}

We provide a formal local analysis of how the adaptive gate suppresses sensitivity to perturbations in the nominal refinement step size on low-alignment inputs.

We know that the TILR update is
\[
h_t
=
\tilde h_t
+
\eta_t \Pi d_t,
\]
where
\[
\eta_t
=
\min\!\left(\eta_{\mathrm{base}}\gamma_t^2,1\right),
\qquad
\gamma_t
=
\frac{\|\Pi d_t\|}{\|d_t\|+\epsilon}.
\]

We first consider the uncapped regime,
$\eta_{\mathrm{base}}\gamma_t^2 < 1$. Substituting the gated step size gives
\[
h_t
=
\tilde h_t
+
\eta_{\mathrm{base}}
\left(
\frac{\|\Pi d_t\|}{\|d_t\|+\epsilon}
\right)^2
\Pi d_t.
\]

We study the local, one-step sensitivity to a perturbation
\[
\eta_{\mathrm{base}}
\rightarrow
\eta_{\mathrm{base}}+\Delta\eta,
\]
holding $\tilde h_t$, $d_t$, and $\Pi$ fixed.

\paragraph{Unconstrained refinement.}
For the standard refinement update
\[
h_t^{\mathrm{ref}}
=
\tilde h_t+\eta_{\mathrm{base}}d_t,
\]
the induced perturbation is
\[
\Delta h_t^{\mathrm{ref}}
=
\Delta\eta\,d_t,
\]
with magnitude
\[
\|\Delta h_t^{\mathrm{ref}}\|
=
|\Delta\eta|\,\|d_t\|.
\]
Thus, the local sensitivity scales linearly with the magnitude of the full
contrastive signal.

\paragraph{TILR update sensitivity.}
For TILR in the uncapped regime, the same perturbation yields
\[
\Delta h_t^{\mathrm{TILR}}
=
\Delta\eta
\left(
\frac{\|\Pi d_t\|}{\|d_t\|+\epsilon}
\right)^2
\Pi d_t,
\]
and therefore
\[
\|\Delta h_t^{\mathrm{TILR}}\|
=
|\Delta\eta|
\frac{\|\Pi d_t\|^3}
     {(\|d_t\|+\epsilon)^2}.
\]

Relative to unconstrained refinement,
\[
\frac{\|\Delta h_t^{\mathrm{TILR}}\|}
     {\|\Delta h_t^{\mathrm{ref}}\|}
=
\gamma_t^2
\frac{\|\Pi d_t\|}{\|d_t\|}.
\]
Since $\Pi$ is an orthogonal projection,
$\|\Pi d_t\|\leq\|d_t\|$, and hence
\[
\frac{\|\Delta h_t^{\mathrm{TILR}}\|}
     {\|\Delta h_t^{\mathrm{ref}}\|}
\leq
\gamma_t^2.
\]
Equivalently,
\[
\|\Delta h_t^{\mathrm{TILR}}\|
\leq
\gamma_t^2
\|\Delta h_t^{\mathrm{ref}}\|
\leq
\|\Delta h_t^{\mathrm{ref}}\|.
\]

Thus, low-alignment inputs exhibit strongly suppressed local sensitivity to
the nominal step size. In particular, for fixed $\|d_t\|$ and small
$\epsilon$, the sensitivity scales cubically with the magnitude of the
projected component:
\[
\|\Delta h_t^{\mathrm{TILR}}\|
=
O\!\left(
\frac{\|\Pi d_t\|^3}{\|d_t\|^2}
\right).
\]
Projection diminishes components outside the learned refinement subspace,
while the squared alignment gate further suppresses corrections whose
projected component is weak. Whereas in the capped regime, $\eta_{\mathrm{base}}\gamma_t^2>1$, the effective step size is locally constant at $\eta_t=1$. Therefore, sufficiently small
perturbations in $\eta_{\mathrm{base}}$ that do not cross the cap boundary
produce no change in the update. This analysis is local and mechanistic: it holds the current state and contrastive direction fixed and characterizes the one-step response to
perturbations in $\eta_{\mathrm{base}}$. This proof does not establish a global trajectory-level sensitivity or generalization guarantee.
\vspace{-0.3cm}
\section{Calibration Set Construction}
\label{app:calibration}

For each benchmark, we draw a calibration set $\mathcal{C} =
\{x_1, \ldots, x_N\}$ uniformly at random from the training split,
with $N{=}200$ and a fixed random seed. None of our benchmarks
provides an official validation split, so calibration is drawn from
training and held disjoint from the test set reported in
Section~\ref{sec:results}.
The procedure requires only forward passes and no ground-truth
labels: for each $x_i$, we run the Coconut backbone with residual
blending, pass each intermediate state $\tilde{h}^t_i$ through both
reference models, and form the contrastive differences
$\delta^t_i = h^t_{\text{good},i} - h^t_{\text{bad},i}$. Stacking
all $NT = 600$ differences and computing a truncated SVD yields the
invariant subspace.
We use $N{=}200$ as a fixed default across all benchmarks rather
than tuning the calibration size per dataset. As shown in
Section~\ref{sec:results}, the resulting contrastive spectra are
concentrated in their leading directions, although the rank required
to explain $90\%$ of the variance varies across datasets
($r_{0.90}=38$--$171$). Total calibration cost is $NT$ reference
forward passes plus the SVD, completing in under 30 seconds on a
single H100. The cost is incurred once per benchmark and reference
pair and is amortized across subsequent inference.
\vspace{-0.3cm}
\section{Hyperparameter Sensitivity}
\label{app:rq2_hyperparam}

We evaluate the sensitivity of refinement methods to the hyperparameters
$(\alpha, \eta)$ controlling residual blending and contrastive update
strength. The goal is to assess whether TILR provides a more stable operating
region than unconstrained refinement. For each dataset, we evaluate accuracy
over a grid of $(\alpha, \eta)$ values and report: (i)~the best accuracy
within the grid, (ii)~the worst-case accuracy in the local neighborhood of
the best point, and (iii)~the sensitivity gap between the best and worst
performance.
Table~\ref{tab:rq2_sensitivity} shows that TILR substantially reduces
hyperparameter sensitivity across all three evaluated benchmarks. On GSM8K,
the sensitivity gap decreases from $0.055$ to $0.003$, while the worst-case
accuracy increases from $0.219$ to $0.275$. Similar improvements are observed
on GSM-Plus, where the gap decreases from $0.041$ to $0.004$, and on SVAMP,
where it decreases from $0.097$ to $0.006$. TILR also improves the best
accuracy on all three benchmarks, from $0.274$ to $0.279$ on GSM8K,
$0.136$ to $0.137$ on GSM-Plus, and $0.315$ to $0.337$ on SVAMP.
These results show that TILR maintains strong performance across a broader
local hyperparameter region rather than relying on a narrow operating point.
This behavior is consistent with the adaptive gate suppressing the effective
update magnitude for low-alignment corrections, thereby reducing sensitivity
to the nominal refinement step size.

\begin{table}[t]
\centering
\caption{\textbf{Hyperparameter sensitivity} 
Each entry reports best accuracy, worst-case accuracy, and sensitivity gap over the non-step hyperparameter grid.
Higher is better for Best and Worst; lower is better for Gap.}
\label{tab:rq2_sensitivity}
\small
\setlength{\tabcolsep}{4pt}
\renewcommand{\arraystretch}{1.0}
\begin{threeparttable}
\begin{adjustbox}{max width=\textwidth}
\begin{tabular}{lccc ccc}
\toprule
\rowcolor{headerblue}
Dataset & \multicolumn{3}{c}{\textbf{Refinement}} & \multicolumn{3}{c}{\textbf{TILR}} \\
\rowcolor{headerblue}
 & Best $\uparrow$ & Worst $\uparrow$ & Gap $\downarrow$
 & Best $\uparrow$ & Worst $\uparrow$ & Gap $\downarrow$ \\
\midrule
GSM8K        & 0.274 & 0.219 & 0.055 & \best{0.279} & \best{0.275} & \best{0.003} \\
GSM-PLUS     & 0.136 & 0.095 & 0.041 & \best{0.137} & \best{0.133} & \best{0.004} \\
SVAMP        & 0.315 & 0.218 & 0.097 & \best{0.337} & \best{0.331} & \best{0.006} \\  
\bottomrule
\end{tabular}
\end{adjustbox}
\end{threeparttable}
\end{table}

\vspace{-0.3cm}
\section{Ablation Study}
\label{app:rq6}

We ablate the components of TILR on GSM8K-Aug
(Table~\ref{tab:ablation}). Starting from the residual-blend backbone
($0.2561$), unconstrained correction improves accuracy to $0.2691$.
Adding subspace projection increases accuracy to $0.2760$, while adaptive
gating alone achieves $0.2737$. Combining both components yields the
strongest performance at $0.2767$, indicating that projection provides the
larger individual gain while gating provides an additional improvement.
We further examine the update design and choice of projection subspace.
Linear damping achieves $0.2752$, while removing the step-size cap reduces
accuracy to $0.2721$. Replacing the learned subspace with its orthogonal
complement, input-PCA directions, or a rank-matched random basis yields
$0.2667$, $0.2335$, and $0.2314$, respectively, all below TILR. These
ablations show that performance depends not only on restricting the update,
but on using the learned refinement directions together with adaptive gating.


\begin{table}[t]
\centering
\caption{Ablation Study on GSM8K-Aug.}
\label{tab:ablation}
\small
\setlength{\tabcolsep}{4pt}
\renewcommand{\arraystretch}{1.0}
\begin{tabular}{lc}
\toprule
\rowcolor{headerblue}
\textbf{Variant} & \textbf{Accuracy} \\
\midrule
Residual blend only ($\eta = 0$, = backbone)   & 0.2561\,$\pm$\,0.014 \\
Correction only (no $\Pi$, no gate)            & 0.2691\,$\pm$\,0.012 \\
$+$ Projection only                            & 0.2760\,$\pm$\,0.010 \\
$+$ Gating only                                & 0.2737\,$\pm$\,0.012 \\
\best{TILR(Projection $+$ Gating)}                                    & \best{0.2767\,$\pm$\,0.012} \\
\midrule
\multicolumn{2}{l}{\emph{Step and damping}} \\
Linear damping ($\eta\gamma$)                  & 0.2752\,$\pm$\,0.013 \\
Uncapped step (no $\min(\cdot,1)$)             & 0.2721\,$\pm$\,0.012 \\
Relative step ($\eta\lVert\tilde h\rVert$)     & 0.2663\,$\pm$\,0.011 \\
\midrule
\multicolumn{2}{l}{\emph{Substituted projectors} (same rank, scale and gate references)} \\
Complement projection $(I-\Pi)$                & 0.2667\,$\pm$\,0.012 \\
Rank-matched input PCA                         & 0.2335\,$\pm$\,0.012 \\
Random rank-171 basis                          & 0.2314\,$\pm$\,0.012 \\
\bottomrule
\end{tabular}
\end{table}
\vspace{-0.3cm}
\section{Cross-Architecture Generalization}
\label{app:cross_arch}

We evaluate TILR on two additional backbones,
Qwen2.5-Math-1.5B~\citep{yang2024qwen25mathtechnicalreportmathematical}
and Llama3.2-1B-Instruct~\citep{grattafiori2024llama3herdmodels},
to test whether the proposed refinement mechanism generalizes beyond the
GPT-2-based models used in the main experiments.
Tables~\ref{tab:rq1-qwen} and~\ref{tab:rq1-llama} show that TILR consistently
improves over unconstrained Refinement across both architectures and all three
evaluated benchmarks. On Qwen2.5-Math-1.5B, TILR improves accuracy by
$+2.4$, $+3.3$, and $+2.8$ percentage points on GSM8K, SVAMP, and GSM-Plus,
respectively. On Llama3.2-1B-Instruct, the corresponding gains are $+1.0$,
$+2.3$, and $+1.0$ points, with comparable generation lengths to the other
latent reasoning methods.
We further evaluate whether the input-invariance improvements generalize across
architectures. As shown in Table~\ref{tab:cross_arch_invariance}, TILR improves
answer agreement (M3) and reduces trajectory variance (M4) relative to
unconstrained Refinement in all six backbone--benchmark combinations. On
Qwen2.5-Math-1.5B, mean answer agreement improves by $0.073$, with trajectory
variance reduced by $3.2$--$27.3\%$ across datasets. On
Llama3.2-1B-Instruct, mean answer agreement improves by $0.030$, with
trajectory-variance reductions of $10.7$--$14.3\%$.
Together with the GPT-2 results, these experiments indicate that both the
accuracy and input-invariance benefits of TILR extend across backbone
architectures and model scales, although the magnitude of the improvement
varies across models and datasets.
\begin{table}[t]
\centering
\caption{Exact-match accuracy (\%) and average generation tokens per inference on three benchmarks for \textbf{Qwen2.5-Math-1.5B} with LoRA (rank 16) and stepwise internalization curriculum.}
\label{tab:rq1-qwen}
\small
\setlength{\tabcolsep}{4pt}
\renewcommand{\arraystretch}{1.15}

\begin{tabular}{lcccccc}
\toprule
\rowcolor{headerblue}
& \multicolumn{2}{c}{\textbf{GSM8K}}
& \multicolumn{2}{c}{\textbf{SVAMP}}
& \multicolumn{2}{c}{\textbf{GSM-Plus}} \\
\rowcolor{headerblue}
\textbf{Method} & \textbf{Acc.} & \textbf{Tok.}
                & \textbf{Acc.} & \textbf{Tok.}
                & \textbf{Acc.} & \textbf{Tok.} \\
\midrule
No-CoT   & $35.4\!\pm\!0.5$ & $3$
         & $32.6\!\pm\!0.6$ & $3$
         & $27.5\!\pm\!0.7$ & $3$ \\
CoT      & $\mathbf{63.5\!\pm\!0.4}$ & $105$ 
         & $47.4\!\pm\!0.4$ & $42$
         & $\mathbf{43.7\!\pm\!0.5}$ & $108$ \\
\midrule
Coconut  & $56.2\!\pm\!0.8$ & $9$ 
         & $51.0\!\pm\!0.9$ & $4$
         & $38.4\!\pm\!1.1$ & $9$ \\
Refinement & $\underline{58.1\!\pm\!0.6}$ & $9$
           & $\underline{52.6\!\pm\!0.7}$ & $4$
           & $\underline{40.2\!\pm\!0.7}$ & $9$ \\
AdaAnchor  & $48.5\!\pm\!1.0$ & $7$
           & $43.8\!\pm\!1.1$ & $4$
           & $34.7\!\pm\!0.9$ & $7$ \\
\midrule
\textbf{TILR (ours)} & \best{$60.5\!\pm\!0.5$} & $9$
                     & \best{$55.9\!\pm\!0.5$} & $4$
                     & \best{$43.0\!\pm\!0.7$} & $9$ \\
\bottomrule

\end{tabular}
\end{table}

\begin{table}[t]
\centering
\caption{Exact-match accuracy (\%) and average generation tokens per inference on three benchmarks for \textbf{Llama3.2-1B-Instruct}.}
\label{tab:rq1-llama}
\small
\setlength{\tabcolsep}{4pt}
\renewcommand{\arraystretch}{1.15}

\begin{tabular}{lcccccc}
\toprule
\rowcolor{headerblue}
& \multicolumn{2}{c}{\textbf{GSM8K}}
& \multicolumn{2}{c}{\textbf{SVAMP}}
& \multicolumn{2}{c}{\textbf{GSM-Plus}} \\
\rowcolor{headerblue}
\textbf{Method} & \textbf{Acc.} & \textbf{Tok.}
                & \textbf{Acc.} & \textbf{Tok.}
                & \textbf{Acc.} & \textbf{Tok.} \\
\midrule
No-CoT   & $30.1\!\pm\!0.3$ & $3$
         & $27.1\!\pm\!0.3$ & $3$
         & $26.1\!\pm\!0.4$ & $3$ \\
CoT      & $\mathbf{60.5\!\pm\!0.4}$ & $105$ 
         & $41.2\!\pm\!0.5$ & $42$
         & $\mathbf{41.5\!\pm\!0.5}$ & $100$ \\
\midrule
Coconut  & $50.1\!\pm\!0.5$ & $9$ 
         & $41.2\!\pm\!0.6$ & $5$
         & $39.1\!\pm\!0.8$ & $7$ \\
Refinement & $\underline{52.5\!\pm\!0.2}$ & $8$
           & $45.6\!\pm\!1.1$ & $4$
           & $\underline{40.1\!\pm\!0.5}$ & $9$ \\
AdaAnchor  & $44.5\!\pm\!1.2$ & $7$
           & \underline{$46.4\!\pm\!0.4$} & $4$
           & $28.2\!\pm\!0.3$ & $7$ \\
\midrule
\textbf{TILR (ours)} & \best{$53.5\!\pm\!0.3$} & $9$
                     & \best{$47.9\!\pm\!0.5$} & $5$
                     & \best{$41.1\!\pm\!0.7$} & $9$ \\
\bottomrule

\end{tabular}
\end{table}

\begin{table}[t]
\centering
\caption{Cross-architecture input invariance under semantically equivalent
reformulations. M3 measures answer agreement ($\uparrow$), and M4 measures
latent trajectory variance ($\downarrow$). Best results within each backbone
and benchmark are shown in bold.}
\label{tab:cross_arch_invariance}
\small
\setlength{\tabcolsep}{4pt}
\renewcommand{\arraystretch}{1.15}

\begin{tabular}{llcccccc}
\toprule
\rowcolor{headerblue}
& & \multicolumn{3}{c}{\textbf{M3: Answer Agreement} $\uparrow$}
& \multicolumn{3}{c}{\textbf{M4: Trajectory Variance} $\downarrow$} \\

\rowcolor{headerblue}
\textbf{Backbone} & \textbf{Dataset}
& \textbf{Coconut}
& \textbf{Refinement}
& \textbf{TILR}
& \textbf{Coconut}
& \textbf{Refinement}
& \textbf{TILR} \\
\midrule

\multirow{3}{*}{\shortstack[l]{\textbf{Qwen2.5-1.5B}}}
& GSM8K
& 0.61 & 0.64 & \best{0.68}
& 0.26 & 0.26 & \best{0.21} \\

& SVAMP
& 0.51 & 0.55 & \best{0.67}
& 0.33 & 0.31 & \best{0.30} \\

& GSM-Plus
& 0.58 & 0.55 & \best{0.61}
& 0.34 & 0.33 & \best{0.24} \\

\midrule

\multirow{3}{*}{\shortstack[l]{\textbf{Llama3.2-1B-Instruct}}}
& GSM8K
& 0.55 & 0.55 & \best{0.57}
& 0.29 & 0.28 & \best{0.25} \\

& SVAMP
& 0.44 & 0.47 & \best{0.51}
& 0.28 & 0.28 & \best{0.24} \\

& GSM-Plus
& 0.51 & 0.52 & \best{0.55}
& 0.22 & 0.22 & \best{0.19} \\

\bottomrule
\end{tabular}
\end{table}
\vspace{-0.3cm}
\section{Broader Impact}

Trajectory Invariant Latent Refinement (TILR) is a training-free inference-time
approach designed to improve the stability of latent reasoning in language
models. By restricting refinement to an empirically identified refinement
subspace and selectively applying updates based on signal alignment, TILR
reduces sensitivity to paraphrasing and reference-checkpoint variations without
requiring retraining. In practice, this leads to more predictable reasoning
behavior under semantically equivalent inputs, which can improve the reliability
and auditability of LLM systems in settings where consistency matters, such as
educational applications, scientific assistance, and decision-support
workflows. Because the method relies only on forward passes and a calibration
SVD for each reference configuration, it also offers a relatively
compute-efficient path toward robustness improvements.
At the same time, TILR does not address the fundamental problem of model
correctness. The method improves invariance and consistency, not factual
accuracy or calibration. If a model exhibits systematic errors or flawed
reasoning, TILR may make those failures more stable and repeatable rather than
correcting them. Although the adaptive gate suppresses unreliable corrections
and recovers the uncorrected backbone as the alignment score approaches zero,
it does not eliminate biases or inaccuracies inherited from the underlying
checkpoints. More broadly, reducing the cost of reliable inference may
accelerate deployment of LLM systems in high-stakes environments before
adequate safeguards, evaluation standards, or verification mechanisms are in
place.
For these reasons, we view TILR as a robustness enhancement rather than a
standalone trustworthiness solution. Downstream deployments should combine it
with task-specific verification, uncertainty estimation, and human oversight
where appropriate, and should avoid treating improved input invariance as
evidence of correctness or safety.
\vspace{-0.3cm}
\section{Limitations}
\label{app:limitations}

TILR improves the stability of latent reasoning, but several limitations
remain. The learned refinement subspace is empirical rather than theoretically
grounded: we do not establish that its directions correspond to causally
invariant reasoning features, and its quality depends on the calibration data
and reference pair. The geometry of the learned subspace also varies across
datasets, and the relationship between this variation and downstream
robustness remains incompletely understood.
TILR improves invariance, not correctness. If reference checkpoints encode
systematic biases or errors, constraining refinement to stable directions does
not necessarily correct them, and consistent trajectories should not be
interpreted as evidence of factual accuracy. The evaluation is also limited
in scale and domain: the main experiments use GPT-2-class backbones, with
Qwen2.5-Math-1.5B and Llama3.2-1B-Instruct evaluated in the appendix, and the
benchmarks focus on mathematical and deductive reasoning. Whether similar
spectrally concentrated refinement structure persists in substantially larger
models or domains such as tool use, planning, and interactive agents remains
open.
Finally, although TILR requires no additional training, it requires calibration
forward passes and a truncated SVD for each reference configuration. How to
maintain stable refinement directions under distribution shift without
repeated recalibration remains an open problem.


\end{document}